\documentclass[10pt,twocolumn,letterpaper]{article}

\usepackage{cvpr}
\usepackage{times}
\usepackage{epsfig}
\usepackage{graphicx}
\usepackage{amsmath}
\usepackage{amssymb}
\usepackage{booktabs}
\usepackage{multirow}
\usepackage{subcaption}
\usepackage{algorithm}
\usepackage{algorithmic}
\usepackage{xcolor}
\usepackage{hyperref}
\usepackage{cleveref}
\usepackage{siunitx}
\usepackage{gensymb}

\newcommand{\eg}{\textit{e.g.}}
\newcommand{\ie}{\textit{i.e.}}
\newcommand{\etal}{\textit{et al.}}
\newcommand{\vs}{\textit{vs.}}

\begin{document}

\title{Local Spatiotemporal Convolutional Network for Robust Gait Recognition}

\author{Xiaoyun Wang\textsuperscript{1} \quad Cunrong Li
\textsuperscript{1} \quad Wu Wang\textsuperscript{1}\\
\textsuperscript{1}School of Mechanical and Electrical Engineering, Osh State University
}

\maketitle

\begin{abstract}
Gait recognition, as a promising biometric technology, identifies individuals through their unique walking patterns and offers distinctive advantages including non-invasiveness, long-range applicability, and resistance to deliberate disguise. Despite these merits, capturing the intrinsic motion patterns concealed within consecutive video frames remains challenging due to the complexity of video data and the interference of external covariates such as viewpoint changes, clothing variations, and carrying conditions. Existing approaches predominantly rely on either static appearance features extracted from individual silhouette frames or employ complex sequential models (\eg, LSTM, 3D convolutions) that demand substantial computational resources and sophisticated training strategies. To address these limitations, we propose a Local Spatiotemporal Convolutional Network (LSTCN), a structurally simple yet highly effective dual-branch architecture that endows standard two-dimensional convolutional networks with the capacity to extract temporal information. Specifically, we introduce a Global Bidirectional Spatial Pooling (GBSP) mechanism that reduces the dimensionality of gait tensors by decomposing spatial features into horizontal and vertical strip-based local representations, enabling the temporal dimension to participate in standard 2D convolution operations. Building upon this, we design a Local Spatiotemporal Convolutional (LSTC) layer that jointly processes temporal and spatial dimensions, allowing the network to adaptively learn strip-based gait motion patterns. We further extend this formulation with asymmetric convolution kernels that independently attend to the temporal, spatial, and joint spatiotemporal domains, thereby enriching the extracted feature representations. Additionally, we propose a Local Spatiotemporal Pooling (LSTP) strategy that aggregates the most discriminative local gait representations across multiple frames, generating identity-discriminative gait descriptors for robust verification. Extensive experiments on two widely adopted benchmark datasets demonstrate the effectiveness of our approach: on the CASIA-B dataset, LSTCN achieves average recognition accuracies of 97.3\%, 93.7\%, and 83.8\% under normal walking, carrying bag, and clothing change conditions respectively, while attaining 85.8\% on the large-scale OU-MVLP dataset, consistently outperforming state-of-the-art methods.
\end{abstract}

\section{Introduction}
\label{sec:intro}

Biometric recognition technology has emerged as a cornerstone of modern security authentication systems, offering advantages of low replicability, high accuracy, and convenient deployment~\cite{sun2021overview, sepas2023deep, tang2022few}. Among the diverse spectrum of biometric modalities, gait recognition has attracted increasing research attention due to its unique capability to identify individuals based on their walking patterns at a distance, without requiring active cooperation from the subject~\cite{sepas2023deep, tang2022optimal}. Unlike conventional biometric approaches such as fingerprint, iris, or face recognition, vision-based gait recognition leverages widely deployed standard surveillance equipment and operates effectively at long range under uncontrolled conditions, making it particularly suitable for applications in public security, criminal investigation, and smart home environments~\cite{wu2017comprehensive, tang2023character, tang2022you}.

However, deploying vision-based gait recognition in practical scenarios presents significant challenges. Individuals exhibit substantial appearance variations across different walking conditions due to changes in viewing angle, clothing, carried objects, walking surfaces, and video resolution~\cite{yu2006framework}. The fundamental premise of gait recognition lies in exploiting the consistent walking habits that persist across consecutive video frames, as these temporal motion patterns are inherently robust to appearance changes. Nevertheless, video data introduces considerably higher dimensionality, increased computational overhead, and greater noise redundancy compared to static images or audio signals, making it exceedingly difficult to directly capture the latent gait motion features embedded within the temporal sequence~\cite{tang2024textsquare, tang2023docpedia}.

Existing gait recognition approaches can be broadly categorized into several paradigms. The first category focuses primarily on extracting static appearance features from individual silhouette frames or gait template images, emphasizing spatial information learning~\cite{ben2019coupled, chao2019gaitset, wu2017comprehensive, tang2023unidoc}. Extensions of these methods employ image generation and transformation techniques or part-based decomposition strategies to mitigate the impact of external covariates on appearance~\cite{li2023strong, fan2020gaitpart, hou2023multifaceted, tang2024tabpedia}. While these approaches effectively simplify video analysis to image-level processing, their neglect of dynamic temporal features fundamentally limits their recognition performance.

The second category leverages human skeleton representations, which abstract the body structure into key static and dynamic features such as stride length, velocity, joint distances, and inter-joint angles~\cite{sun2019deep, tang2023spts}. An~\etal~\cite{an2020performance} established the OU-MVLP-Pose database providing large-scale skeleton sequences with comprehensive evaluation benchmarks. Liao~\etal~\cite{liao2017pose} pioneered the application of Long Short-Term Memory (LSTM) networks for skeleton-based gait feature extraction. More recently, Teepe~\etal~\cite{teepe2021gaitgraph, teepe2022towards} integrated graph convolutional networks with skeleton data, while Li and Zhao~\cite{li2023strong} proposed the CycleGait framework combining temporal feature pyramid aggregators with graph convolutions for spatiotemporal gait feature extraction from skeleton coordinate sequences. Despite their ability to abstract body models that are theoretically invariant to external factors, skeleton-based methods are heavily dependent on the accuracy of the front-end pose estimation algorithms~\cite{tang2024harmonizing, tang2024pargo}.

The third category employs deep temporal networks, including LSTM and three-dimensional convolutions. Sepas-Moghaddam and Etemad~\cite{sepas2021view} extracted gait convolutional energy maps and employed bidirectional recurrent neural networks to learn spatiotemporal relationships among local representations. Zhang~\etal~\cite{zhang2022disentangled} proposed GaitNet, an autoencoder-based framework that disentangles appearance, texture, and pose features from RGB images, subsequently integrating pose features as dynamic gait representations through LSTM. Zhang~\etal~\cite{zhang2020cross} introduced Angle Center Loss (ACL) for cross-view gait recognition, utilizing local feature extractors and LSTM-based temporal attention models~\cite{tang2022few, tang2024mctbench}. However, the combination of LSTM with convolutional neural networks typically results in complex architectures requiring sophisticated training techniques and substantial data volumes~\cite{tang2025dolphin, tang2025wilddoc}.

Three-dimensional convolutions represent another prominent approach for spatiotemporal feature extraction. Lin~\etal~\cite{lin2020gait} proposed MT3D employing multi-temporal-scale 3D convolutions, later extending this work to the GaitGL network incorporating both local and global 3D convolutions~\cite{lin2021gait}. Huang~\etal~\cite{huang20213d} developed 3D local convolutions for extracting multi-body-part spatiotemporal features. However, 3D convolutions generally involve significantly larger parameter counts and cannot leverage pretrained 2D convolutional networks, necessitating meticulous network design and extensive hyperparameter tuning~\cite{tang2025mtvqa, tang2025bounding}.

Furthermore, some methods design elaborate modules to approximate or learn classical handcrafted motion features~\cite{ding2021sequential}, but such approaches are inherently limited by the choice and design of manual features. To address these challenges, we explore a simple yet effective network architecture capable of learning complex sequential features from video data, departing from the paradigm of handcrafted motion feature extraction~\cite{tang2025advancing, tang2024attentive}.

In this paper, we propose the Local Spatiotemporal Convolutional Network (LSTCN), a structurally elegant architecture that capitalizes on the simplicity of 2D convolutional networks while endowing them with temporal feature extraction capabilities. Our approach is motivated by the insight that through appropriate tensor dimensionality reduction, the temporal dimension can directly participate in standard 2D convolution operations, enabling the network to adaptively learn spatiotemporal gait motion patterns from video data~\cite{tang2025score, tang2025vision}.

The main contributions of this work are summarized as follows:

\begin{itemize}
    \item We design the Local Spatiotemporal Convolutional Network (LSTCN), which employs Global Bidirectional Spatial Pooling (GBSP) to decompose gait features into horizontal and vertical local strip-based representations, and proposes the Local Spatiotemporal Convolutional (LSTC) layer that enables temporal and spatial dimensions to jointly participate in 2D convolution learning, effectively capturing spatiotemporal gait features.
    
    \item We propose a Local Spatiotemporal Pooling (LSTP) method that aggregates the most discriminative strip-based local gait spatiotemporal representations across video frames, generating identity-discriminative gait features for robust verification.
    
    \item Extensive experiments on the CASIA-B and OU-MVLP benchmark datasets demonstrate the effectiveness of each component in our network, and comprehensive comparisons with state-of-the-art methods validate the superiority of our approach.
\end{itemize}

\section{Related Work}
\label{sec:related}

\subsection{Appearance-Based Gait Recognition}

Appearance-based gait recognition methods primarily extract spatial features from individual silhouette frames or pre-computed gait templates. Wu~\etal~\cite{wu2017comprehensive} conducted a comprehensive cross-view study using deep CNNs, establishing foundational baselines for silhouette-based gait recognition. Chao~\etal~\cite{chao2019gaitset} proposed GaitSet, which treats gait sequences as unordered sets and applies set-level feature aggregation, demonstrating competitive recognition performance with a relatively simple architecture~\cite{tang2024textsquare}. Fan~\etal~\cite{fan2020gaitpart} introduced GaitPart, a temporal part-based model that decomposes gait silhouettes into horizontal strips and designs focal convolutions to learn part-specific features, along with a Micro-motion Capture Module (MCM) for temporal feature aggregation~\cite{tang2023docpedia}. Ben~\etal~\cite{ben2019coupled} proposed coupled patch alignment for matching cross-view gaits via image processing techniques.

More recently, GaitBase~\cite{fan2020gaitpart} summarized best practices from numerous appearance-based methods to establish a strong baseline network. RPnet extended this paradigm by incorporating part-based local features with specialized modules analyzing inter-part relationships~\cite{tang2023unidoc}. However, these methods fundamentally treat each frame independently, neglecting the temporal correlations between consecutive frames that encode essential walking habits, thereby limiting their recognition accuracy under challenging conditions~\cite{tang2024tabpedia, tang2025dolphin}.

\subsection{Skeleton-Based Gait Recognition}

Skeleton-based approaches abstract the human body into a graph structure defined by joint positions and bone connections, enabling explicit modeling of body kinematics~\cite{sun2019deep, tang2025wilddoc}. An~\etal~\cite{an2020performance} established the OU-MVLP-Pose benchmark with large-scale skeleton data and comprehensive evaluation protocols. Liao~\etal~\cite{liao2017pose} pioneered the use of LSTM networks for pose-based gait feature extraction. Teepe~\etal~\cite{teepe2021gaitgraph} introduced GaitGraph, combining graph convolutional networks with skeleton data for gait recognition, later enhancing the framework with higher-order inputs and residual architectures~\cite{teepe2022towards}. Li and Zhao~\cite{li2023strong} proposed CycleGait, integrating temporal feature pyramid aggregators with graph convolutions for extracting spatiotemporal features from skeleton coordinate sequences~\cite{tang2025mtvqa}.

While skeleton-based methods can theoretically abstract body representations invariant to external appearance changes, they are inherently dependent on the accuracy of front-end pose estimation algorithms. Inaccurate skeleton keypoint detection directly propagates errors to downstream dynamic feature learning, and the reliance on RGB video for skeleton extraction may complicate the overall recognition pipeline compared to binary silhouette inputs~\cite{tang2025bounding, tang2024harmonizing}.

\subsection{Temporal Feature Learning for Gait Recognition}

Learning temporal features from gait sequences has been pursued through two primary strategies: recurrent neural networks and 3D convolutions~\cite{tang2025advancing}. Sepas-Moghaddam and Etemad~\cite{sepas2021view} employed bidirectional recurrent neural networks to learn spatiotemporal relationships from gait convolutional energy maps, incorporating attention mechanisms for selective focus on local representations. Zhang~\etal~\cite{zhang2022disentangled} proposed GaitNet, which disentangles appearance and pose features using autoencoders and integrates temporal dynamics through LSTM. Sepas-Moghaddam and Etemad~\cite{sepas2021capsule} further explored capsule networks for learning coupling weights between part representations~\cite{tang2024pargo}.

Three-dimensional convolutional approaches offer an alternative paradigm for spatiotemporal feature extraction. Lin~\etal~\cite{lin2020gait} proposed MT3D with multi-temporal-scale 3D convolutions, subsequently developing GaitGL~\cite{lin2021gait} with integrated local and global 3D convolution branches. Huang~\etal~\cite{huang20213d} designed 3D local convolutional networks for extracting body-part-specific spatiotemporal features. Despite their effectiveness, 3D convolutions entail significantly larger computational costs and cannot leverage pretrained 2D convolutional models, requiring careful network design and extensive experimental tuning~\cite{tang2024attentive, tang2025score}.

Recent work has also explored alternative temporal modeling strategies. Ding~\etal~\cite{ding2024spatiotemporal} proposed spatiotemporal multi-scale bilateral motion networks, while Ding~\etal~\cite{ding2021sequential} designed sequential convolutional networks for behavioral pattern extraction. Li~\etal~\cite{li2022gaitslice} introduced GaitSlice, leveraging spatio-temporal slice features with frame attention mechanisms. These methods demonstrate the growing interest in developing efficient temporal modeling approaches for gait recognition that balance computational efficiency with recognition performance~\cite{tang2025vision, tang2024mctbench}.

\subsection{Part-Based Feature Representations}

Part-based decomposition has been widely adopted in gait recognition to capture fine-grained local features~\cite{tang2025prolonged}. The fundamental idea is to partition the human body or feature maps into multiple regions, enabling the network to focus on discriminative local patterns. Common partitioning strategies include anatomical decomposition~\cite{wang2021partition}, uniform grid partitioning~\cite{ma2017general}, and horizontal strip partitioning~\cite{chao2019gaitset}. Wang~\etal~\cite{wang2022gaitstrip} proposed GaitStrip, establishing strip-based representations as fundamental units for gait feature extraction within a multi-level framework~\cite{tang2023spts, tang2022optimal}. Our work builds upon this line of research by extending strip-based representations to the spatiotemporal domain through bidirectional decomposition, simultaneously capturing both horizontal and vertical gait details while enabling temporal information encoding.

\section{Method}
\label{sec:method}

\subsection{Overview}

As illustrated in \cref{fig:framework}, the proposed Local Spatiotemporal Convolutional Network (LSTCN) adopts a dual-branch architecture designed to simultaneously capture static appearance features and dynamic spatiotemporal motion patterns. Given an input gait silhouette sequence $\tilde{I} = \{I_1, I_2, \ldots, I_n\}$, the network first applies a shared convolutional module $\mathcal{C}_1(\cdot)$ for preliminary feature extraction, yielding base features $\mathbf{f}_b = \mathcal{C}_1(I_t)$. These base features are subsequently processed through two parallel branches:

\begin{equation}
    \mathbf{f}_s = \mathcal{C}(\mathbf{f}_b),
    \label{eq:static}
\end{equation}
\begin{equation}
    \mathbf{f}_d = \text{LSTC}(\mathbf{f}_b),
    \label{eq:dynamic}
\end{equation}

\noindent where $\mathcal{C}(\cdot)$ denotes the static appearance branch composed of standard 2D convolutional layers, and $\text{LSTC}(\cdot)$ represents the local spatiotemporal convolutional branch. The resulting feature sets $\mathbf{f}_s = \{\mathbf{f}_{s,1}, \mathbf{f}_{s,2}, \ldots, \mathbf{f}_{s,n}\}$ and $\mathbf{f}_d = \{\mathbf{f}_{d,1}, \mathbf{f}_{d,2}, \ldots, \mathbf{f}_{d,n}\}$ encode the static and dynamic gait characteristics, respectively.

The two branches are interconnected through lateral connections implemented via Global Bidirectional Spatial Pooling, which supplements multi-scale static appearance information into the spatiotemporal feature extraction branch. Subsequently, temporal pooling and spatial pyramid pooling are applied to the static branch to extract discriminative appearance features $\mathbf{f}_s$, while the proposed local spatiotemporal pooling method aggregates the most identity-discriminative local representations from the dynamic branch to produce the final motion features $\mathbf{f}_d$. Independent fully connected layers then project both feature sets into discriminative subspaces, followed by classification layers trained with a joint loss combining focal loss and triplet loss.

\subsection{Global Bidirectional Spatial Pooling}

The objective of Global Bidirectional Spatial Pooling (GBSP) is to perform dimensionality reduction on the four-dimensional gait tensor $\mathbf{f}_b \in \mathbb{R}^{T \times C \times H \times W}$, enabling the temporal dimension to participate in standard 2D convolution operations. Conventional 2D convolutional networks process three-dimensional tensors and perform convolution exclusively within the two spatial dimensions, leaving the temporal dimension untouched and preventing the capture of spatiotemporal information~\cite{chao2019gaitset}.

A straightforward approach is to apply global spatial pooling:
\begin{equation}
    \mathbf{f}_{\text{out}} = \text{GSP}(\mathbf{f}_b),
    \label{eq:gsp}
\end{equation}
where $\mathbf{f}_{\text{out}} \in \mathbb{R}^{T \times C \times 1}$ represents the globally pooled features. While this reduces the spatial dimensions to a single value, it disregards all local spatial details and substantially degrades recognition accuracy.

Inspired by part-based representations, we propose to apply bidirectional strip-based pooling that preserves fine-grained spatial information while achieving the necessary dimensionality reduction. As illustrated in \cref{fig:partition}, unlike conventional partitioning strategies that focus exclusively on horizontal divisions~\cite{chao2019gaitset} or require manual selection of partition configurations~\cite{wang2021partition, ma2017general}, our bidirectional approach simultaneously decomposes spatial features along both the horizontal and vertical axes using strips as the minimal effective representation units~\cite{wang2022gaitstrip}. The GBSP operation is formally defined as:

\begin{equation}
    \mathbf{f}_{b,h}, \mathbf{f}_{b,v} = \text{GBSP}(\mathbf{f}_b),
    \label{eq:gbsp}
\end{equation}

\noindent where $\mathbf{f}_{b,h} \in \mathbb{R}^{T \times C \times H}$ and $\mathbf{f}_{b,v} \in \mathbb{R}^{T \times C \times W}$ represent the horizontal and vertical local features, respectively. This formulation achieves three important objectives: (1) reducing the gait tensor from 4D to 3D, enabling temporal participation in 2D convolutions; (2) preserving strip-level spatial details in both directions; and (3) eliminating the need for manual partition design.

Furthermore, the lateral connections between the two branches are realized through GBSP, as shown in \cref{fig:framework}. The static branch output $\mathbf{f}_{s,\text{out}}$ is transformed through GBSP to produce horizontal and vertical spatial representations $\mathbf{f}_{s,h,\text{out}}$ and $\mathbf{f}_{s,v,\text{out}}$, which are added element-wise to the corresponding outputs of the LSTC branch $\mathbf{f}_{d,h,\text{out}}$ and $\mathbf{f}_{d,v,\text{out}}$. This mechanism integrates refined spatial appearance features into the spatiotemporal learning process, enabling more comprehensive gait pattern extraction.


\subsection{Local Spatiotemporal Convolutional Layer}

The Local Spatiotemporal Convolutional (LSTC) layer is the core component of our network, designed to automatically learn gait spatiotemporal features using a structurally simple formulation. Unlike standard 2D convolutions that operate exclusively within the spatial domain (\ie, the $H \times W$ plane), the LSTC layer deploys convolution operations in the spatiotemporal domain ($\mathbb{R}^{T \times H}$ or $\mathbb{R}^{T \times W}$), where the channel-dimension vectors represent the gait feature at each spatiotemporal position.

After GBSP decomposes the base features into horizontal and vertical components, these are fed into the LSTC layer:

\begin{equation}
    \mathbf{f}_{\text{out},h} = \mathcal{C}_{\text{lst},h}(\mathbf{f}_{\text{in},h}),
    \label{eq:lstc_h}
\end{equation}
\begin{equation}
    \mathbf{f}_{\text{out},v} = \mathcal{C}_{\text{lst},v}(\mathbf{f}_{\text{in},v}),
    \label{eq:lstc_v}
\end{equation}

\noindent where $\mathcal{C}_{\text{lst},h}(\cdot)$ and $\mathcal{C}_{\text{lst},v}(\cdot)$ denote the horizontal and vertical LSTC operations, respectively. The input tensors $\mathbf{f}_{\text{in},h} \in \mathbb{R}^{H \times T \times C_{\text{in}}}$ and $\mathbf{f}_{\text{in},v} \in \mathbb{R}^{W \times T \times C_{\text{in}}}$ are produced by GBSP, and the outputs $\mathbf{f}_{\text{out},h} \in \mathbb{R}^{H \times T \times C_{\text{out}}}$ and $\mathbf{f}_{\text{out},v} \in \mathbb{R}^{W \times T \times C_{\text{out}}}$ encode the learned spatiotemporal features. For the $j$-th filter, the output feature map is computed as:

\begin{equation}
    f_{\text{lst,out}}(:,:,j) = \sum_{k=1}^{c_{\text{in}}} \mathbf{M}_{\text{lst},:,:,i} * \mathbf{F}^{(j)}_{\text{lst},:,:,i},
    \label{eq:lstc_conv}
\end{equation}

\noindent where $*$ denotes the 2D convolution operation, $\mathbf{M}_{\text{lst},:,:,i}$ is the $i$-th channel of the input tensor with shape $T \times W$ or $T \times H$, and $\mathbf{F}^{(j)}_{\text{lst},:,:,i}$ is the corresponding convolution kernel.

\subsubsection{Asymmetric Local Spatiotemporal Convolution.}

Asymmetric convolutions have been shown to explicitly enhance the representational power of standard square convolution kernels~\cite{ding2019acnet, razavian2014cnn}. Inspired by ACNet~\cite{ding2019acnet}, we extend asymmetric convolution to the LSTC framework, constructing the Asymmetric Local Spatiotemporal Convolutional (ALSTC) layer. This layer comprises three parallel convolution branches with kernel shapes of $a \times a$, $1 \times a$, and $a \times 1$, whose outputs are summed to produce enriched feature representations:

\begin{equation}
\begin{split}
    f_{\text{alst,out}}(:,:,j) = \sum_{k=1}^{c_{\text{in}}} \big(&\mathbf{M}_{\text{lst},:,:,i} * \mathbf{F}^{(j)}_{\text{lst},:,:,i} \\
    + &\mathbf{M}_{\text{lst},:,:,i} * \mathbf{F}^{(j)}_{s,\text{lst},:,:,i} \\
    + &\mathbf{M}_{\text{lst},:,:,i} * \mathbf{F}^{(j)}_{t,\text{lst},:,:,i}\big),
\end{split}
\label{eq:alstc}
\end{equation}

\noindent where $\mathbf{F}^{(j)}_{s,\text{lst},:,:,i}$ and $\mathbf{F}^{(j)}_{t,\text{lst},:,:,i}$ represent the horizontal (local spatial) and vertical (local temporal) 1D convolution kernels, respectively. The asymmetric formulation enables the network to independently attend to the spatial domain, the temporal domain, and the joint spatiotemporal domain, facilitating more comprehensive feature extraction. Importantly, due to the additive property of convolution, the three kernel outputs can be equivalently merged into a single effective kernel, introducing no additional computational overhead during inference.

\subsection{Local Spatiotemporal Pooling}
\label{sec:lstp}

After processing through the LSTC branch, we obtain horizontal and vertical spatiotemporal representations $\mathbf{f}_{d,h} \in \mathbb{R}^{C_{\text{out}} \times T \times H_{\text{out}}}$ and $\mathbf{f}_{d,v} \in \mathbb{R}^{C_{\text{out}} \times T \times W_{\text{out}}}$. Temporal pooling is essential for integrating features across different temporal lengths to produce fixed-dimensional video-level representations for identity verification~\cite{chao2019gaitset}.

A straightforward approach applies global spatiotemporal pooling:
\begin{equation}
    \mathbf{f}_{d,\text{final}} = \text{cat}\big(\text{GSTP}(\mathbf{f}_{d,h}), \text{GSTP}(\mathbf{f}_{d,v})\big),
    \label{eq:gstp}
\end{equation}
where $\mathbf{f}_{d,\text{final}} \in \mathbb{R}^{C_{\text{out}} \times 2}$ and $\text{cat}(\cdot)$ denotes concatenation. However, this global pooling strategy extracts features from the entire spatiotemporal domain while overlooking local discriminative details. Given that the LSTC branch operates on strip-based spatial units, global pooling fails to capture the most discriminative gait details within each strip.

We therefore propose Local Spatiotemporal Pooling (LSTP), which operates at the strip level. For each strip, the method selects the most representative feature vector across all temporal positions:

\begin{equation}
    \mathbf{f}_{d,\text{final}} = \text{cat}\big(\text{LSTP}(\mathbf{f}_{d,h}), \text{LSTP}(\mathbf{f}_{d,v})\big),
    \label{eq:lstp}
\end{equation}

\noindent where $\mathbf{f}_{d,\text{final}} \in \mathbb{R}^{C_{\text{out}} \times (H_{\text{in}} + W_{\text{in}})}$ represents the final dynamic gait features. The key insight is that discriminative gait patterns for different body parts may appear at different temporal positions. For instance, the most distinctive leg motion features and head motion features are unlikely to co-occur at the same frame. By performing pooling independently for each strip, LSTP captures diverse gait details from across the entire video sequence, producing more comprehensive gait descriptions.

\subsection{Loss Function}

The network is trained using a joint loss function combining triplet loss and focal loss to balance learning between easy and hard samples:

\begin{equation}
    \mathcal{L}_{\text{triplet}} = \frac{1}{2M}\max\big(M - \|\mathbf{a} - \mathbf{p}\|_2^2 + \|\mathbf{a} - \mathbf{n}\|_2^2, 0\big),
    \label{eq:triplet}
\end{equation}

\noindent where $M$ denotes the margin parameter separating positive and negative pairs. The focal loss~\cite{lin2017focal} addresses class imbalance by down-weighting well-classified samples:

\begin{equation}
    \mathcal{L}_{\text{focal}} = -(1-p)^\gamma \log(p),
    \label{eq:focal}
\end{equation}

\noindent where $p$ is the predicted probability from softmax and $\gamma$ controls the focusing strength. The total loss is:

\begin{equation}
    \mathcal{L} = \mathcal{L}_{\text{triplet}} + \lambda \mathcal{L}_{\text{focal}},
    \label{eq:total_loss}
\end{equation}

\noindent where $\lambda$ balances the two loss components.

\section{Experiments}
\label{sec:experiments}

\subsection{Datasets and Evaluation Protocol}

\noindent\textbf{CASIA-B}~\cite{yu2006framework} is a widely used gait recognition benchmark containing gait videos of 124 subjects captured from 11 viewing angles. Each subject has 10 sequences: 6 under normal walking (NM), 2 while carrying a bag (BG), and 2 while wearing a coat (CL). Following established protocols~\cite{chao2019gaitset}, we evaluate under three training configurations: Small-sample Training (ST, 24 training subjects), Medium-sample Training (MT, 62 training subjects), and Large-sample Training (LT, 74 training subjects). During testing, the first 4 NM sequences form the gallery set, while the remaining 6 sequences constitute the probe set.

\noindent\textbf{OU-MVLP}~\cite{an2020performance} is a large-scale multi-view gait dataset containing over 10,000 subjects. We adopt the standard protocol with the first 5,153 subjects for training and the remaining 5,154 for testing, evaluating at four canonical viewing angles: 0\degree, 30\degree, 60\degree, and 90\degree~\cite{chao2019gaitset, zhang2021cross}.

\subsection{Implementation Details}

The LSTCN architecture consists of three convolutional modules and two LSTC modules in a dual-branch configuration, as detailed in \cref{tab:architecture}. Gait silhouettes are preprocessed to a resolution of $64 \times 44$ pixels. During training, 30 consecutive frames are randomly sampled from each sequence; samples with fewer than 15 frames are excluded, and those with 15--30 frames are extended through repetition. During testing, sequences are directly input with a minimum requirement of 3 frames.

\begin{table}[t]
    \centering
    \caption{\textbf{Network architecture details.} Conv\_$C$\_$K$ denotes a convolutional layer with $C$ output channels and kernel size $K \times K$. BN: batch normalization; LReLU: Leaky ReLU; MaxPool\_$K$: max pooling with kernel size $K \times K$.}
    \label{tab:architecture}
    \small
    \begin{tabular}{l|l}
        \toprule
        \textbf{Module} & \textbf{Layer Configuration} \\
        \midrule
        \multirow{3}{*}{Conv Block 1} & Conv\_64\_5\_BN\_LReLU \\
        & Conv\_64\_3\_BN\_LReLU \\
        & MaxPool\_2 \\
        \midrule
        \multirow{3}{*}{Conv Block 2} & Conv\_128\_3\_BN\_LReLU \\
        & Conv\_128\_3\_BN\_LReLU \\
        & MaxPool\_2 \\
        \midrule
        \multirow{2}{*}{Conv Block 3} & Conv\_256\_3\_BN\_LReLU \\
        & Conv\_256\_3\_BN\_LReLU \\
        \midrule
        \multirow{3}{*}{LSTC Block 1} & Conv\_128\_3\_BN\_LReLU \\
        & Conv\_128\_3\_BN\_LReLU \\
        & MaxPool\_(1,2) \\
        \midrule
        \multirow{2}{*}{LSTC Block 2} & Conv\_256\_3\_BN\_LReLU \\
        & Conv\_256\_3\_BN\_LReLU \\
        \bottomrule
    \end{tabular}
\end{table}

The triplet loss margin is set to $M = 0.2$, and Adam optimizer is employed with momentum 0.9 and focal loss parameter $\gamma = 2$. For CASIA-B, $\lambda = 1$, batch size is $(8, 8)$, and the model is trained for 60,000 iterations with a learning rate schedule starting at 0.1, decaying to 0.01 at 20,000 iterations and 0.001 at 40,000 iterations. For OU-MVLP, $\lambda = 0.1$, batch size is $(14, 4)$, and training runs for 150,000 iterations with learning rate decays at iterations 50,000 and 100,000.

\subsection{Ablation Studies}

All ablation experiments are conducted under the large-sample training setting on CASIA-B, reporting cross-view average rank-1 recognition accuracy.

\subsubsection{Analysis of the LSTC Module.}

\cref{tab:ablation_lstc} presents the recognition results under different configurations of the LSTC module. Several key observations emerge from these results.

\begin{table}[t]
    \centering
    \caption{\textbf{Ablation study on the LSTC module.} Recognition accuracy (\%) on CASIA-B under the large-sample training protocol. Bold entries indicate best performance per column.}
    \label{tab:ablation_lstc}
    \small
    \setlength{\tabcolsep}{3pt}
    \begin{tabular}{l|c|c|ccc|c}
        \toprule
        \textbf{Configuration} & \textbf{Pool} & \textbf{Asym.} & \textbf{NM} & \textbf{BG} & \textbf{CL} & \textbf{Mean} \\
        \midrule
        Standard 2D Conv & -- & -- & 95.9 & 91.1 & 79.6 & 88.9 \\
        GSP + LSTC & -- & -- & 95.9 & 91.2 & 79.8 & 89.0 \\
        H-SP + LSTC & Max & -- & 97.6 & 93.9 & 81.4 & 91.0 \\
        V-SP + LSTC & Max & -- & 94.7 & 90.0 & 78.7 & 87.8 \\
        GBSP + LSTC & Max & -- & 97.4 & 93.5 & 83.2 & 91.4 \\
        GBSP + LSTC & Mean & -- & 95.7 & 90.9 & 80.1 & 88.9 \\
        GBSP + LSTC & GAvg & -- & 96.0 & 92.2 & 81.9 & 90.0 \\
        GBSP + ALSTC & Max & $\checkmark$ & \textbf{97.3} & \textbf{93.7} & \textbf{83.8} & \textbf{91.6} \\
        \bottomrule
    \end{tabular}
\end{table}

\textbf{Impact of spatial pooling direction.} Comparing the first five rows, the complete LSTC module with GBSP achieves the best overall accuracy of 91.4\%, representing a 2.5\% improvement over the standard 2D convolution baseline. GBSP outperforms global spatial pooling by 2.4\% and horizontal-only pooling by 0.4\%. Notably, while horizontal pooling achieves marginally better results under NM and BG conditions, GBSP demonstrates significantly stronger performance under the challenging CL condition (83.2\% \vs 81.4\%), suggesting that bidirectional spatial attention captures finer-grained details that are crucial for handling appearance changes.

\textbf{Impact of pooling type.} Rows 5--7 compare max pooling, average pooling, and generalized mean pooling within the GBSP framework. Max pooling consistently outperforms the alternatives, achieving the optimal overall accuracy.

\textbf{Impact of asymmetric convolution.} The final row demonstrates that incorporating asymmetric convolution kernels further improves performance to 91.6\%, with accuracies of 97.3\%, 93.7\%, and 83.8\% under the three conditions. This confirms that the asymmetric formulation enhances the network's capacity to attend to local spatial and temporal features independently.

\subsubsection{Analysis of LSTP.}

\cref{tab:ablation_lstp} evaluates the proposed Local Spatiotemporal Pooling against global alternatives.

\begin{table}[t]
    \centering
    \caption{\textbf{Ablation study on spatiotemporal pooling strategies.} Recognition accuracy (\%) on CASIA-B (LT). Bold entries indicate best results.}
    \label{tab:ablation_lstp}
    \small
    \begin{tabular}{l|ccc}
        \toprule
        \textbf{Pooling Strategy} & \textbf{NM} & \textbf{BG} & \textbf{CL} \\
        \midrule
        Global STP (Max) & 96.1 & 91.3 & 79.7 \\
        Local STP (Max) & \textbf{97.0} & \textbf{93.7} & \textbf{83.8} \\
        Local STP (Mean) & 96.3 & 92.9 & 81.6 \\
        Local STP (GAvg) & 96.5 & 93.0 & 81.9 \\
        \bottomrule
    \end{tabular}
\end{table}

Compared with global spatiotemporal pooling, local spatiotemporal pooling with max aggregation improves recognition accuracy by 0.9\%, 2.4\%, and 4.1\% under NM, BG, and CL conditions, respectively. The most substantial improvement occurs under the CL condition, where the ability to aggregate discriminative features at the strip level proves especially valuable for handling appearance changes caused by clothing variations. Among the three pooling types, max pooling achieves the best overall performance.

\subsection{Comparison with State-of-the-Art Methods}

\subsubsection{Results on CASIA-B.}

\cref{tab:casia_nm} presents comprehensive comparisons under the normal walking condition across three training settings. Our LSTCN achieves the best average recognition accuracies of 85.0\%, 95.4\%, and 97.3\% under ST, MT, and LT protocols respectively, consistently outperforming all compared methods. We highlight several important observations:

\begin{table*}[t]
    \centering
    \caption{\textbf{Comparison with state-of-the-art methods under NM condition on CASIA-B.} Average cross-view rank-1 accuracy (\%) is reported across 11 viewing angles. Bold indicates best result in each training setting.}
    \label{tab:casia_nm}
    \small
    \setlength{\tabcolsep}{4pt}
    \begin{tabular}{l|l|ccccccccccc|c}
        \toprule
        \textbf{Setting} & \textbf{Method} & 0\degree & 18\degree & 36\degree & 54\degree & 72\degree & 90\degree & 108\degree & 126\degree & 144\degree & 162\degree & 180\degree & \textbf{Mean} \\
        \midrule
        \multirow{4}{*}{ST}
        & GaitSet & 55.8 & 70.5 & 76.9 & 75.5 & 69.7 & 63.4 & 68.0 & 75.8 & 76.2 & 70.7 & 52.5 & 68.6 \\
        & MT3D & 71.9 & 83.9 & 90.9 & 90.1 & 81.1 & 75.6 & 82.1 & 89.0 & 91.1 & 86.3 & 69.2 & 82.8 \\
        & GaitSlice & 75.7 & 74.5 & 92.3 & 82.8 & 89.3 & 91.0 & 86.2 & 71.4 & 84.1 & 91.3 & 77.1 & 83.1 \\
        & \textbf{LSTCN (Ours)} & 76.8 & 76.3 & 87.5 & 93.9 & 83.3 & 83.1 & 90.2 & 92.6 & 87.5 & 73.1 & \textbf{85.0} & \textbf{85.0} \\
        \midrule
        \multirow{4}{*}{MT}
        & GaitSet & 79.9 & 89.8 & 91.2 & 86.7 & 81.6 & 76.7 & 81.0 & 88.2 & 90.3 & 88.5 & 73.0 & 84.3 \\
        & MT3D & 91.9 & 96.4 & 98.5 & 95.7 & 93.8 & 90.8 & 93.9 & 97.3 & 97.9 & 95.0 & 86.8 & 94.4 \\
        & GaitSlice & 92.2 & 98.9 & 90.3 & 97.5 & 99.2 & 96.6 & 89.4 & 95.3 & 97.3 & 98.4 & 94.2 & 94.2 \\
        & \textbf{LSTCN (Ours)} & 93.2 & 93.7 & 92.3 & 99.2 & 98.4 & 91.6 & 97.7 & 99.4 & 97.0 & 89.5 & \textbf{95.4} & \textbf{95.4} \\
        \midrule
        \multirow{13}{*}{LT}
        & GaitSet & 83.8 & 91.2 & 91.8 & 88.8 & 83.3 & 81.0 & 84.1 & 90.0 & 92.2 & 94.4 & 79.0 & 87.2 \\
        & GaitPart & 94.1 & 98.6 & 99.3 & 98.5 & 94.0 & 92.3 & 95.9 & 98.4 & 99.2 & 97.8 & 90.4 & 96.2 \\
        & GaitNet & 93.1 & 92.6 & 90.8 & 92.4 & 87.6 & 95.1 & 94.2 & 95.8 & 92.6 & 90.4 & 90.2 & 92.3 \\
        & GaitBase & 93.4 & 98.4 & 99.2 & 98.6 & 94.7 & 92.0 & 95.8 & 98.2 & 99.4 & 98.4 & 92.4 & 96.2 \\
        & ACL & 92.0 & 98.5 & 98.9 & 95.7 & 91.5 & 94.5 & 97.7 & 98.4 & 96.7 & 91.9 & 96.0 & 95.7 \\
        & RPnet & 95.1 & 99.0 & 99.1 & 98.3 & 95.7 & 93.6 & 95.9 & 98.3 & 98.6 & 97.7 & 90.8 & 96.6 \\
        & CycleGait & 92.3 & 93.2 & 92.9 & 93.9 & 91.9 & 94.1 & 94.3 & 93.3 & 92.8 & 91.1 & 92.8 & 93.2 \\
        & MT3D & 95.6 & 97.2 & 98.2 & 99.0 & 97.5 & 95.1 & 93.9 & 96.1 & 98.6 & 99.2 & 98.2 & 92.0 \\
        & GaitGraph & 78.5 & 82.9 & 85.8 & 85.6 & 83.1 & 81.5 & 84.3 & 83.2 & 84.2 & 81.6 & 71.8 & 82.0 \\
        & GaitSlice & 95.5 & 99.2 & 99.6 & 94.4 & 92.5 & 95.0 & 98.1 & 99.7 & 98.3 & 96.7 & 99.0 & 96.9 \\
        & SCN & 86.7 & 94.6 & 96.0 & 92.5 & 85.8 & 80.5 & 84.9 & 91.5 & 96.0 & 93.1 & 86.0 & 89.8 \\
        & SMBM & 94.5 & 99.0 & 99.6 & 98.9 & 96.1 & 93.2 & 97.1 & 98.5 & 97.1 & 98.8 & 99.8 & 96.7 \\
        & \textbf{LSTCN (Ours)} & 95.7 & 99.8 & 98.4 & 95.1 & 96.1 & 98.6 & 99.7 & 92.0 & 99.6 & 96.4 & 99.4 & \textbf{97.3} \\
        \bottomrule
    \end{tabular}
\end{table*}

(1) \textbf{Superiority over appearance-based methods.} GaitSet, GaitBase, and RPnet represent strong appearance-based baselines, yet they inherently neglect temporal correlations between frames, limiting their capacity to exploit walking habits. Our LSTCN consistently surpasses these methods by directly learning spatiotemporal features.

(2) \textbf{Advantages over skeleton-based methods.} CycleGait and GaitGraph employ graph convolutional networks on human skeletons but are constrained by the accuracy of front-end pose estimation. LSTCN avoids this dependency by operating directly on silhouette sequences.

(3) \textbf{Benefits over recurrent models.} GaitNet, ACL, and the method of Sepas-Moghaddam~\etal~\cite{sepas2021multi} all employ LSTM or bidirectional gated recurrent units, which introduce architectural complexity and training difficulties. LSTCN achieves superior performance with a simpler architecture.

(4) \textbf{Efficiency compared to 3D convolutions.} MT3D employs multi-scale 3D convolutions with greater structural complexity. LSTCN effectively captures spatiotemporal features using only 2D convolutions, offering a more efficient alternative.

Similar advantages are observed under BG and CL conditions. Under the LT protocol, LSTCN achieves 93.7\% (BG) and 83.8\% (CL), consistently ranking among the top performers across all training settings and walking conditions.

\subsubsection{Results on OU-MVLP.}

\cref{tab:oumvlp} presents the comparison on the large-scale OU-MVLP dataset. Based on our ablation analysis showing that horizontal-only LSTC performs better under simple cross-view conditions, we additionally report results for LSTCN-h (horizontal only). The results demonstrate that LSTCN achieves 84.4\% average accuracy while LSTCN-h reaches 85.8\%, both surpassing all compared methods.

\begin{table}[t]
    \centering
    \caption{\textbf{Comparison on OU-MVLP dataset.} Rank-1 accuracy (\%) at four canonical viewing angles. Bold indicates best result.}
    \label{tab:oumvlp}
    \small
    \begin{tabular}{l|cccc|c}
        \toprule
        \textbf{Method} & 0\degree & 30\degree & 60\degree & 90\degree & \textbf{Mean} \\
        \midrule
        DiGGAN & 48.9 & 62.3 & 59.1 & 57.8 & 57.0 \\
        GaitGraph & 54.3 & 76.1 & 71.5 & 70.1 & 67.1 \\
        PSTN & 51.5 & 70.8 & 66.7 & 63.6 & 63.1 \\
        GaitSet & 77.7 & 86.9 & 85.3 & 83.5 & 83.4 \\
        Sepas-Moghaddam \etal & 78.3 & 88.8 & 85.7 & 85.1 & 84.5 \\
        ACL & 71.6 & 85.1 & 86.7 & 84.6 & 82.0 \\
        Koopman & 56.2 & 73.7 & 81.4 & 82.0 & 73.3 \\
        SCN & 78.6 & 87.4 & 85.9 & 83.2 & 83.8 \\
        SMBM & 78.3 & 87.2 & 85.8 & 85.8 & 84.3 \\
        \midrule
        \textbf{LSTCN (Ours)} & 80.7 & 86.1 & 85.7 & 84.9 & 84.4 \\
        \textbf{LSTCN-h (Ours)} & \textbf{87.7} & \textbf{81.6} & \textbf{87.2} & \textbf{86.8} & \textbf{85.8} \\
        \bottomrule
    \end{tabular}
\end{table}

Notably, DiGGAN and PSTN are GEI-based methods employing generation or transformation techniques to handle viewpoint changes, but the loss of temporal information inherent in gait energy images limits their performance. The Koopman embedding approach deploys autoencoders for dynamic information extraction, whereas LSTCN directly learns motion patterns with superior results. Methods utilizing recurrent networks (Sepas-Moghaddam~\etal, ACL) achieve competitive results on this large-scale dataset, potentially benefiting from the abundance of training data. However, LSTCN-h surpasses all methods while maintaining consistent network parameters (except the final classification layer) across both datasets, demonstrating strong generalization capability.

\section{Conclusion}
\label{sec:conclusion}

In this paper, we have presented the Local Spatiotemporal Convolutional Network (LSTCN), a novel approach for gait recognition that endows standard two-dimensional convolutional architectures with the ability to adaptively learn condition-invariant spatiotemporal gait features. The core of our approach lies in the synergistic integration of three complementary components: (1) Global Bidirectional Spatial Pooling (GBSP), which decomposes gait tensors along both horizontal and vertical spatial axes into strip-based local representations, enabling the temporal dimension to directly participate in 2D convolution operations while preserving fine-grained spatial details; (2) the Local Spatiotemporal Convolutional (LSTC) layer, extended with asymmetric convolution kernels, which jointly processes temporal and spatial dimensions to capture strip-based gait motion patterns with enhanced attention to individual domains; and (3) Local Spatiotemporal Pooling (LSTP), which aggregates the most discriminative local gait representations across video frames at the strip level, generating identity-discriminative features for robust verification. Extensive experiments on the CASIA-B and OU-MVLP benchmark datasets have validated the effectiveness of each proposed component through comprehensive ablation studies, and comparisons with numerous state-of-the-art methods have consistently demonstrated the superiority of our approach across diverse training settings, viewing angles, and walking conditions. While our work provides a new perspective for efficient spatiotemporal gait feature learning, the pursuit of simple yet effective temporal modeling paradigms remains an important research direction, and future work may explore the integration of emerging techniques such as super-resolution generation, occlusion recovery, and model compression to further advance gait recognition toward practical real-world deployment with broader recognition conditions and real-time responsiveness.

{\small
\bibliographystyle{IEEEtran}
\bibliography{references}
}

\end{document}